\documentclass[11pt]{article}

\usepackage[T1]{fontenc}
\usepackage[utf8]{inputenc}
\usepackage{lmodern}
\usepackage{microtype}
\usepackage[a4paper,margin=1in]{geometry}
\usepackage{amsmath,amssymb}
\usepackage{booktabs}
\usepackage{array}
\usepackage{tabularx}
\usepackage{longtable}
\usepackage{graphicx}
\usepackage{caption}
\usepackage{csquotes}
\usepackage{xurl}
\usepackage[hidelinks]{hyperref}
\usepackage[
  style=authoryear-comp,
  backend=biber,
  maxbibnames=99,
  maxcitenames=2,
  uniquelist=false
]{biblatex}

\graphicspath{{figures/}}

\newenvironment{fullwidthtable}[1][htbp]
  {\begin{table}[#1]}
  {\end{table}}

\title{\textbf{When ``Victorian'' Becomes a Prompt:}\\
Literary Periodization as a Generative Constraint in 100 AI-Generated Novels}

\author{
Mehdy Sedaghat Payam\\
\small Independent Scholar\\
\small \href{mailto:mehdyspayam@gmail.com}{mehdyspayam@gmail.com}\\
\small \href{https://orcid.org/0000-0002-9171-2622}{ORCID: 0000-0002-9171-2622}
}

\date{September 2026}

\begin{document}

\maketitle

\begin{abstract}
Generative AI inverts the typical periodization of literary history: the periodizing tag ``Victorian'' can now come first and influence what is written. ``Generative periodization,'' defined and tested here, describes the use of literary-period designations in generating texts. I test this approach on 100 book-length novels produced under Victorian and Zero-Style conditions using GPT, Qwen, and Llama workflows. The Period Alignment Score (PAS), trained on nineteenth-century literature and benchmarked against human Zero-Style prose, assesses alignment using topic-reduced grammatical features. Victorian prompts produce consistent historical-direction shifts in GPT and Qwen, but not robustly in Llama. Victorian-only recalibration and harder comparison corpora preserve the GPT and Qwen effects. Cross-model transfer also shows a shared direction of grammatical change. The measurable target is the broader nineteenth century rather than the Victorian period per se.
\end{abstract}


\bigskip

\section{Introduction}

Literary periods are usually defined retrospectively. Scholars identify recurring historical and literary patterns and use labels such as Romantic, Victorian, modernist, or postwar to organize them. Such categories are neither neutral nor fully stable: as Underwood argues, periodization simplifies differences among works and reflects the history of literary and academic institutions. Yet its basic sequence is familiar: works are produced first and categories are applied afterward. \textcite{underwood2013} traces the institutional history of this contrastive logic, while his later computational work shows that period boundaries can also appear as gradual distributions rather than fixed divisions \parencite{underwood2019}.

Generative artificial intelligence has changed this usual process. Now, before the text is written, the name of a literary period can be placed in the prompt, and this name can affect the form and features of the text from the beginning and guide its production. This change leads to an important question for computational literary studies: when a historical and literary category, such as ``Victorian,'' becomes part of the process of producing a text, what change happens to that category? In other words, how does generative AI interpret this prompt and what kind of text does it generate?

This question is different from whether large language models can imitate the style of a specific writer. Several studies have used methods such as prompt writing, providing examples in the prompt text, measuring stylistic distance, identifying the author, and detailed stylistic comparison to examine how successfully AI can imitate the style of prominent writers. \textcite{bhandarkar2024} showed that available systems cannot reliably reproduce the style of a writer, but they can approximate it to some extent. \textcite{mikros2025} prompted GPT-4o to produce texts in the style of Ernest Hemingway and Mary Shelley and reached a similar result. The generated texts were, in some respects, close to the style of these writers but failed to reproduce the style of the original works fully.

More recent research in computational stylometry has examined this issue in relation to literary and political writers \parencite{alsadhan2026} and review authors \parencite{ishihara2026}. At the same time, \textcite{chen2026}, \textcite{osullivan2025}, and \textcite{sedaghatpayamquinn2026}, respectively analyzing AI-generated essays, short fiction, and novels, show that distinguishable stylistic features can separate human from AI writing. \textcite{wang2025} likewise show that artificial intelligence struggles to reproduce individual styles in less structured formats such as blogs and forums. Across these studies, style can be guided and measured, but generative AI still has difficulty reproducing it fully. A different question therefore remains: what happens when the goal is not to imitate a specific writer but to reproduce a literary category formed across history?

A label like ``Victorian'' goes beyond the linguistic style of one specific writer and refers to a historical and complex collection of concepts including time periods, institutions, genres, cultural memories, important and accepted literary works, later imitations of those works, and critics' descriptions. This is why, when we place the word ``Victorian'' in the prompt of a large language model, it cannot simply turn to one specific text and copy its features. In this context, the model faces the challenge of turning a general cultural concept into textual features and patterns. For this reason, the generated text can help us examine how artificial intelligence turns literary history into something that can be produced in text.

I call this process ``generative periodization,'' which for this paper means a process in which a historical and literary category is turned into an instruction that affects the form of the text before its production. This paper argues that literary-period labels can function as selective generative constraints: they can reproducibly shift formal distributions without fully reproducing the historical category they name. This study does not claim that it can access the model's internal understanding of Victorian literature, nor does it claim that the generated texts are evidence that a model truly ``understands'' a literary period. Instead, it asks a more limited question: when the prompt asks generative AI to generate texts related to a specific literary period, which measurable features change, and how consistently do those changes appear across different large language models?

This study analyzes 100 book-length novels generated by artificial intelligence. These novels were produced using GPT, Qwen, and Llama under two conditions: some with an instruction related to Victorian style and others under contemporary, almost unmarked stylistic conditions that \textcite{quinn2025} has called ``Zero-Style.'' These generated texts are then compared with 205 nineteenth-century human novels, 65 human Zero-Style novels, and 40 modern historical novels written by contemporary writers from the late twentieth and early twenty-first centuries. This paper goes beyond identifying the presence of historical words in the text as a sign of similarity to Victorian literature and uses a measure called the ``Period Alignment Score'' (PAS). This measure compares features such as function words, patterns of parts of speech (PoS), and grammatical features between human and AI-generated texts. To limit the influence of prolific authors, the human-written corpus is author-balanced during cross-validation.

\section{From Literary Periodization to Generative Periodization}

Literary periodization necessarily simplifies heterogeneous texts, writers, genres, institutions, and cultural developments. Nevertheless, labels such as ``Victorian'' remain useful for comparison. \textcite{underwood2013,underwood2019} shows both the historical importance of period contrast and the value of treating literary change probabilistically, through shifting distributions of textual features. Computational literary history extends this change of scale by examining patterns across larger collections rather than only a small number of canonical works \parencite{moretti2005,jockers2013,piper2018,bode2018}. Very frequent, less noticeable features such as function words and grammatical patterns can reveal stylistic and historical variation \parencite{burrows2002,jockers2013,venglarova2024}. \textcite{craiggreatleyhirsch2017} similarly argue that lexical items are more salient but more contingent on topic and situation, whereas frequent function words reveal more persistent stylistic variation.

This distinction matters here because features that make a text immediately recognizable as ``Victorian'' may not necessarily be the same features that computational analysis finds historically diagnostic.

Current scholarship on neo-Victorian literature makes this issue more visible. \textcite{heilmannllewellyn2010} treat neo-Victorianism as both a return to and a renewed engagement with the cultural memory of the Victorian period, and thus as an indirect and mediated way of accessing the nineteenth century. \textcite{hadley2010} similarly emphasizes that contemporary historical fiction constructs ``the Victorians'' from the perspective of the present. This study examines that mediation through a corpus of forty modern historical novels. Computational analysis of these novels shows how closely contemporary human prose can approach nineteenth-century grammatical patterns while remaining a modern historical representation.

This same distinction is also important in evaluating AI-generated Victorian fiction. A generated novel may have clear and recognizable signs of Victorian fiction, but structurally it may remain closer to contemporary prose. Conversely, the text may become closer to nineteenth-century patterns in terms of some less obvious features, such as function words and grammatical patterns, without its tone sounding clearly old or archaic. For this reason, we cannot say ``generative periodization'' has occurred simply because the model has followed the instruction of the prompt. Instead, this study examines three questions separately: does the Victorian instruction cause a change in the generated prose? Are these changes aligned with the differences seen between historical and contemporary human writing? And are similar changes also seen across different models and methods of text generation?

Generative periodization does not turn a literary period into an algorithm; rather, it changes how a literary label functions. A label such as ``Victorian,'' formerly used mainly to categorize and explain existing texts, can now directly shape the production of new texts. Examining and measuring this change allows computational literary studies to turn a familiar literary category into a subject for experimentation. Instead of only asking what features lead us to call a text ``Victorian,'' we can ask what features appear when a text-generation system is asked to produce a Victorian text.

Finally, operationalization is not only a neutral stage that follows conceptual clarification; it is itself part of the theoretical problem. \textcite{pichlerreiter2022} consider operationalization in digital humanities a process that connects theoretical concepts with observable and measurable features in the text. In this study, ``Victorian'' is located on both sides of this process. First, the word is given to the text-generation system as a cultural concept. Then, the generated texts are compared with textual patterns determined on the basis of real historical works. The experiment therefore investigates not only whether the prompt creates a shift in style, but also what remains after a historically contested category becomes computationally operationalized. A mismatch between the prompt label and the historical dimension provides data about the process of operationalization rather than simply constituting measurement error.

The dataset contains 410 novels: 100 AI-generated and 310 human-written. Table~\ref{tab:1} summarizes the groups. These corpora were selected specifically for the purposes of this study and are not intended to represent literary history as a whole; each also plays a different role in the analysis. The historical comparison group contains 205 English-language novels published between 1800 and 1900 by 121 authors. These texts come from the Victorian Novels corpus created by the SSHRC-funded Ciphers of The Times project \parencite{ciphers2022}.

The broad historical human corpus contains novels of varied subgenres and lengths. Their inclusion together does not imply that nineteenth-century fiction is homogeneous; the corpus also reflects selection biases, including which works were digitized and which became part of the canon.

The 65-text human ``Zero-Style'' collection was selected in direct consultation with Justin Quinn. His concept of a kind of simple and globally understandable English, which he develops in \textit{Literature in the Age of Lingua Franca English: The Zero Style} \parencite{quinn2025}, is one of the theoretical foundations of this paper. This collection is not intended to be a representative sample of contemporary literature. The selection criterion was mainly based on excluding works that have very distinctive linguistic features. As a result, texts were selected that use relatively simple and understandable English and make less use of local dialects, historical language, highly experimental styles, or features that are very specific to a particular genre. This collection includes works by 29 authors published between 1982 and 2025. Therefore, it is better understood as a collection for comparison with prose from the 1980s to the contemporary period, rather than as a random and representative sample of present-day fiction.

The modern historical novel comparison corpus includes 40 novels by 30 authors that were published between 1988 and 2023. These works were purposefully selected from modern historical novels whose main story world reconstructs the nineteenth century or returns to that period. This corpus is not used in PAS training and serves only as an interpretive control group. Its purpose is to examine how closely modern historical novels written by humans can approach historical prose in terms of grammatical patterns. These 40 works are presented neither as a complete collection of neo-Victorian literature nor as a statistically representative sample of all historical novels.

There are forty GPT novels, forty Qwen novels, and twenty Llama novels in the AI corpus, equally balanced between the two experimental conditions: the Victorian condition called for nineteenth-century British realist prose, set in a historically bound social world; the Zero-Style condition called for contemporary English with minimal historical, regional, experimental, or genre-based style. The ``Zero-Style'' condition is thus an operational control, not an assertion of the stylelessness of any prose.

There are substantial differences between corpora in terms of total text size. The median text length ranges from approximately 26,500 words in the Llama novels to 178,600 words for the GPT Victorian novels. To prevent these differences from influencing the results, the same target sample is used for every text in all experiments. This sample consists of ten equally spaced windows of 2,000 words each, totaling 20,000 words for every novel. If the text has fewer than 20,000 words, it is divided into ten equal parts. This sampling strategy preserves coverage across the narrative while making feature rates comparable across corpora.

\begin{fullwidthtable}[htbp]
\centering
\begin{tabular}{@{}p{0.28\linewidth}p{0.18\linewidth}rrr@{}}
\toprule
Corpus/workflow & Condition & texts & authors & Median full words \\
\midrule
Human nineteenth-century & Human19C & 205 & 121 & 144,007 \\
Human Zero-Style comparator & HumanZeroStyle & 65 & 29 & 72,912 \\
Modern historical human & ModernHistorical & 40 & 30 & 123,789 \\
GPT & Victorian & 20 & -- & 178,617 \\
GPT & Zero-Style & 20 & -- & 104,545 \\
Qwen3-14B & Victorian & 20 & -- & 52,867 \\
Qwen3-14B & Zero-Style & 20 & -- & 52,715 \\
Llama 3.1 & Victorian & 10 & -- & 27,062 \\
Llama 3.1 & Zero-Style & 10 & -- & 26,135 \\
\bottomrule
\end{tabular}
\caption{Corpus composition and median full-text length.}
\label{tab:1}
\end{fullwidthtable}

\subsection{Generation Workflows and Prompt Conditions}

The three AI subcorpora are not a perfectly balanced 3 $\times$ 2 factorial design experiment. They were generated under different conditions, including differences in interface, decoding controls, segmentation, context handling, and scaffolding. Therefore, ``model'' in this study is shorthand for a model-generation workflow and does not denote a pure architectural effect.

The GPT corpus was interactively created using the ChatGPT interface by ``GPT-5.5 Thinking.'' All forty novels in this corpus were generated chapter by chapter, and the ChatGPT interface provided no information about backend snapshots, seeds, temperatures, top-p, or other decoding parameters. The prompt for the Victorian novels asked for English long-form fiction set in nineteenth-century Britain, with either third-person or omniscient perspective, social observation, varied rhythm, historically realistic institutions, and no modern elements or author imitations. The prompt for Zero-Style novels asked for globally legible English-language novels with minimal regional, historical, experimental, and named-author marking.

For the Qwen collection, the Qwen3-14B model \parencite{yang2025} was used in the Google Colab environment. The model was run with 4-bit NF4 quantization, and computations were carried out with BF16. The model's ``explicit thinking'' capability was also disabled. For text generation, temperature was set to .70, top-p to .80, top-k to 20, and repetition penalty to 1.05. A fixed method was also used for determining the seed. The forty Qwen novels are especially important because they were generated from twenty neutral story ideas. Each idea was generated once under the Victorian condition and once under the Zero-Style condition, and the generation order of the two conditions was balanced. In both conditions, the main story idea was the same. Therefore, this design makes a paired comparison possible in which the primary intended difference is the period-style condition.

For the Llama collection, the meta-llama/Llama-3.1-8B-Instruct checkpoint \parencite{grattafiori2024} was used with 4-bit NF4 quantization. Each novel was generated in twenty chapters, and each chapter was divided into three parts. For text generation, temperature was set to .82, top-p to .92, top-k to 50, and repetition penalty to 1.08. A fixed method was also used for determining the seed, and during the generation process, a compact summary of the story state was carried from one section to the next. The Llama collection is smaller than the other two collections and has only ten novels for each condition. For this reason, its results are treated as a lower-powered replication used primarily to examine the general direction of the results.

Because prompts and generation procedures differ across workflows, the primary comparison is Victorian versus Zero-Style within each workflow. Qwen offers the strongest internal evidence because story premises are held constant across conditions. GPT provides replication under a different, interface-mediated workflow whose backend details are unavailable, while the smaller Llama corpus serves as a lower-powered directional replication.

\subsection{Topic-Reduced Linguistic Representation}

The primary representation is designed to measure historical alignment without rewarding direct references to historical content. Each sampled novel is analyzed with spaCy's English-language processing pipeline. Named-entity recognition is disabled because names of people, places, and organizations are not used in the feature set.

The stored representation includes 286 features:

\begin{itemize}

\item the usage rates of 131 function words;

\item the rates of eleven parts of speech;

\item the rates of 121 consecutive pairs of parts of speech;

\item the rates of eleven selected syntactic dependency relations;

\item and the rates of twelve general grammatical features.

\end{itemize}

All these rates are calculated and standardized per 10,000 alphabetic tokens.

The function-word list includes articles, auxiliary verbs, modal verbs, pronouns, conjunctions, prepositions, determiners, and other frequent grammatical words. Part-of-speech features include eleven broad grammatical categories, while 121 observed POS bigrams capture local grammatical sequencing. Syntactic-dependency and aggregate features include relative and subordinate constructions, coordination, modals, auxiliaries, passive constructions, pronoun groups, demonstratives, conjunctions, prepositions, and contractions such as don't and isn't.

Proper nouns, named entities, semantic embeddings, dates, and lists of words that are directly related to historical topics have been excluded from the analysis. This decision is central to the main research question. A novel generated in the Victorian style should not receive a high historical score only because estates, carriages, priests, or shillings have been mentioned in it. Instead, the main model examines whether the text-generation instruction can change basic linguistic patterns---patterns that show the characteristics of grammar and the use of function words in historical prose. Function words have long been used in stylometry because their frequency, compared with other words, is less dependent on the topic of the text, and consciously controlling them in long texts is difficult \parencite{burrows2002,stamatatos2009}.

\subsection{Contrastive Human-Calibrated Period Alignment Score}

The main PAS measure is built from the outset on a comparison between two groups. This measure is trained using 205 nineteenth-century human novels as the positive group and 65 works of human prose without a specified style (Zero-Style) as the negative group. A standardized logistic regression model with L2 regularization distinguishes these two collections from each other based on 286 linguistic features. The fitted model then calculates the corresponding historical-group score for the remaining texts.

Therefore, a higher PAS means that, in this specific comparison between two human groups, the text has greater similarity to the historical group. But PAS by itself cannot determine whether a text is ``more similar to nineteenth-century works'' or merely ``less similar to prose without a specified style.'' PAS is not the probability that a text was composed in a particular period, a measurable distance between texts, or a context-independent measure for assessing the historical character of a text.

To prevent the unequal influence of authors, two precautionary measures have been taken. First, the five-fold StratifiedGroupKFold method places all the novels of one author in one fold. For example, this method prevents one of Dickens's novels from being placed in the training set and another of his novels in the validation set. Second, the weight of each author during training is inversely proportional to the number of novels that the author has in each period. The total weights of the two groups are also rescaled so that they are equal. Therefore, every author, regardless of the number of their works in the corpus, has the same overall influence on the training of the model.
The model's performance is reported both at the level of weighted novels and at the level of authors. At the author level, before evaluation, out-of-fold scores are averaged across all novels by each author. After cross-validation, the final model with author balancing is trained on the entire human calibration set and is run, without retraining, on all 410 texts, including the modern historical and AI corpora.

Three categories of robustness tests examine the degree of the results' dependence on this comparison. First, the historical positive group is limited to 152 novels published from 1837 to 1900, and pre-Victorian novels are set aside. Second, this analysis tests whether the negative group drives the effect observed in AI texts. To do this, the model was trained again with two alternative groups: all 105 modern human texts, including 65 works without a specified style and 40 modern historical novels; and only those same 40 modern historical novels, as a deliberately difficult negative group. Each alternative is run with both the broad nineteenth-century group and the restricted Victorian group. Third, because logistic-regression probabilities may saturate near 0 or 1, the raw output of the decision function is linearly scaled: zero for the average of human authors without a specified style and one for the average of nineteenth-century authors.

\subsection{Statistical Comparisons}

Within each AI-generation workflow, the distributions of PAS scores in Victorian texts and texts without a specified style are compared using differences between means, Hedges's g measure, Cliff's delta measure, bootstrap confidence intervals, and permutation tests with 5,000 resamples. The confidence intervals use 2,000 bootstrap samples. Because the Llama collection has only ten novels in each condition, interpretation emphasizes effect size and directional consistency rather than statistical significance alone.

The paired Qwen analysis uses the twenty recovered story-premise IDs. In each pair, the Zero-Style PAS score is subtracted from the Victorian PAS score. This analysis reports the paired standardized effect size dz, a sign-flip permutation test, a bootstrap interval for the mean paired change, and the Wilcoxon signed-rank test.

As a supplementary check for differences in full-text length and workflow baselines, a robust OLS model predicts logit-transformed PAS from generation condition, model type, their interaction, and the logarithm of word count. HC3 standard errors are used. Because feature extraction is standardized for length by sampling 20,000 words from each text, total word count is used here as an auxiliary sensitivity variable rather than as the primary correction for differences in text length.

The Victorian-only calibration repeats the tests of each generation process and the paired Qwen test with the same resampling methods. The normalized decision-margin analysis is a method for examining the sensitivity of the results to scale, not an independent classifier. Because this analysis is a rescaling of the same trained decision function, it examines whether the main conclusions depend on probability saturation. This analysis does not provide a separate representation of the features.

\subsection{Cross-Model Period Transfer}

To test whether the Victorian condition creates a similar formal direction across models, a logistic classifier is trained on one workflow to distinguish Victorian from Zero-Style AI novels. This classifier is then applied to another workflow without retraining. For each train-test direction, balanced accuracy, AUC, and MCC are recorded. AUC is the primary measure because it shows whether the learned score puts Victorian texts above Zero-Style texts despite changes in model-specific baselines.
In the second analysis, which is entirely descriptive, the features of each AI model before transfer are standardized according to the overall mean and variance of that same model. This model-centered analysis examines the transferability of the direction of the difference between the two conditions after the baseline differences are removed. This method is transductive because it uses the distribution of the test model's unlabeled features for centering. Therefore, it should be considered a method for understanding the mechanism, not a usable prediction for out-of-domain data.

By bootstrapping the training and test samples, we can evaluate uncertainty in transfer between different models. Qwen resampling is performed on complete matched pairs of story premise and seed; GPT and Llama resampling is performed within each condition. A complementary permutation test breaks the association between the Victorian and Zero-Style labels in the training process while preserving group balance; in Qwen, this is done by randomly swapping labels within matched pairs. The observed raw AUC is the primary cross-model measure, while centered accuracy remains a tool for understanding the mechanism.

\subsection{Author-Balanced Human Period Markers}

The marker analysis examines which linguistic features separate nineteenth-century human prose from contemporary prose and whether the Victorian condition in AI texts also changes those features in the same direction. To prevent prolific authors from having excessive influence on the tests, the mean of each feature is first calculated for the works of each author. Then, the effect size of the difference between the two human periods is calculated based on the author means. A feature is considered a stable period marker when the absolute value of Hedges's g is at least 0.25, the q value adjusted by the Benjamini--Hochberg false-discovery-rate method is below 0.05, and the 95\% bootstrap interval for the mean difference does not include zero.

For each stable marker, the difference between Victorian texts and texts without a specified style is calculated separately in GPT, Qwen, and Llama. This analysis shows how many historical markers present in human writing change in the expected direction in all three generation processes and in at least two processes. To examine whether this number is greater than the amount expected by chance while the relationship among linguistic features is preserved, a permutation test with 10,000 iterations randomly permutes the condition labels within each AI process without shuffling the feature matrix. In Qwen, labels are swapped only within matched story-premise and seed pairs. The resulting null distribution records the number of aligned stable human markers in all three processes, in at least two processes, and in each process separately.

\section{Results}

\subsection{Human Period Classification}

Author balancing does not weaken the historical classifier. At the novel level, where all authors have equal total weight, the balanced accuracy is 0.9784, AUC is 0.9996, and MCC is 0.9578. At the author level, the balanced accuracy is 0.9828, AUC is 0.9997, and MCC is 0.9786 (Table~\ref{tab:2}). These figures demonstrate that the topic-reduced grammatical representation contains a clear period signal that is not simply caused by repeated representation of particular authors.

The final PAS calibration also places the three human corpora in the expected order. Based on the average at the author level, the mean PAS is 0.9981 for nineteenth-century fiction, 0.0020 for contemporary fiction, and 0.4161 for modern historical fiction. Therefore, modern historical fiction occupies an intermediate region between the other two groups rather than approaching either endpoint. This corpus, which was not used in training the model, is important for interpreting the AI results because it shows that contemporary writers can become substantially closer to historical prose in terms of grammatical patterns without being nineteenth-century authors.

Restricting the historical positive group to 152 novels from 1837 to 1900 maintains the major trend (author AUC = .9992): GPT g = 1.299, Qwen g = 1.059, and Llama g = 1.211, and the paired Qwen effect is dz = .764 (sign-flip p = .0001). However, even the 53 non-Victorian novels not included in the above analysis receive scores very close to the historical end of the spectrum. Consequently, the human continuum defined by the model applies more generally to nineteenth-century fiction than specifically to the Victorian era. Changing the negative group yields a further result. When all 105 modern human texts or the subset of 40 modern historical novels serve as the negative group, the GPT effect is still strong (g = 1.11--1.24; all p = .0004), and the Qwen effect is still strong (g = .90--.98; all p = .0004); the paired Qwen effect is still dz = .65--.71 with a sign-flip p = .0001. But the Llama effect falls to g = .10--.34 and becomes non-significant with the new negative group.

These robustness results show that the main GPT and Qwen effects cannot be attributed solely to the intentionally low historical marking of the Zero-Style comparator. By contrast, the Llama result depends more strongly on how the modern comparison group is constructed. The alternative negative classes therefore reduce concern that the primary GPT and Qwen findings are artifacts of the Zero-Style corpus design, although these alternatives are still not fully representative samples of contemporary literary fiction.

\begin{table*}[htbp]
\centering
\begin{tabular}{@{}lrrr@{}}
\toprule
Evaluation unit & Balanced accuracy & AUC & MCC \\
\midrule
Novel level, author-weighted & .9784 & .9996 & .9578 \\
Author level & .9828 & .9997 & .9786 \\
\bottomrule
\end{tabular}
\caption{Author-balanced human period-classification performance.}
\label{tab:2}
\end{table*}

\begin{figure*}[htbp]
\centering
\includegraphics[width=0.85\linewidth]{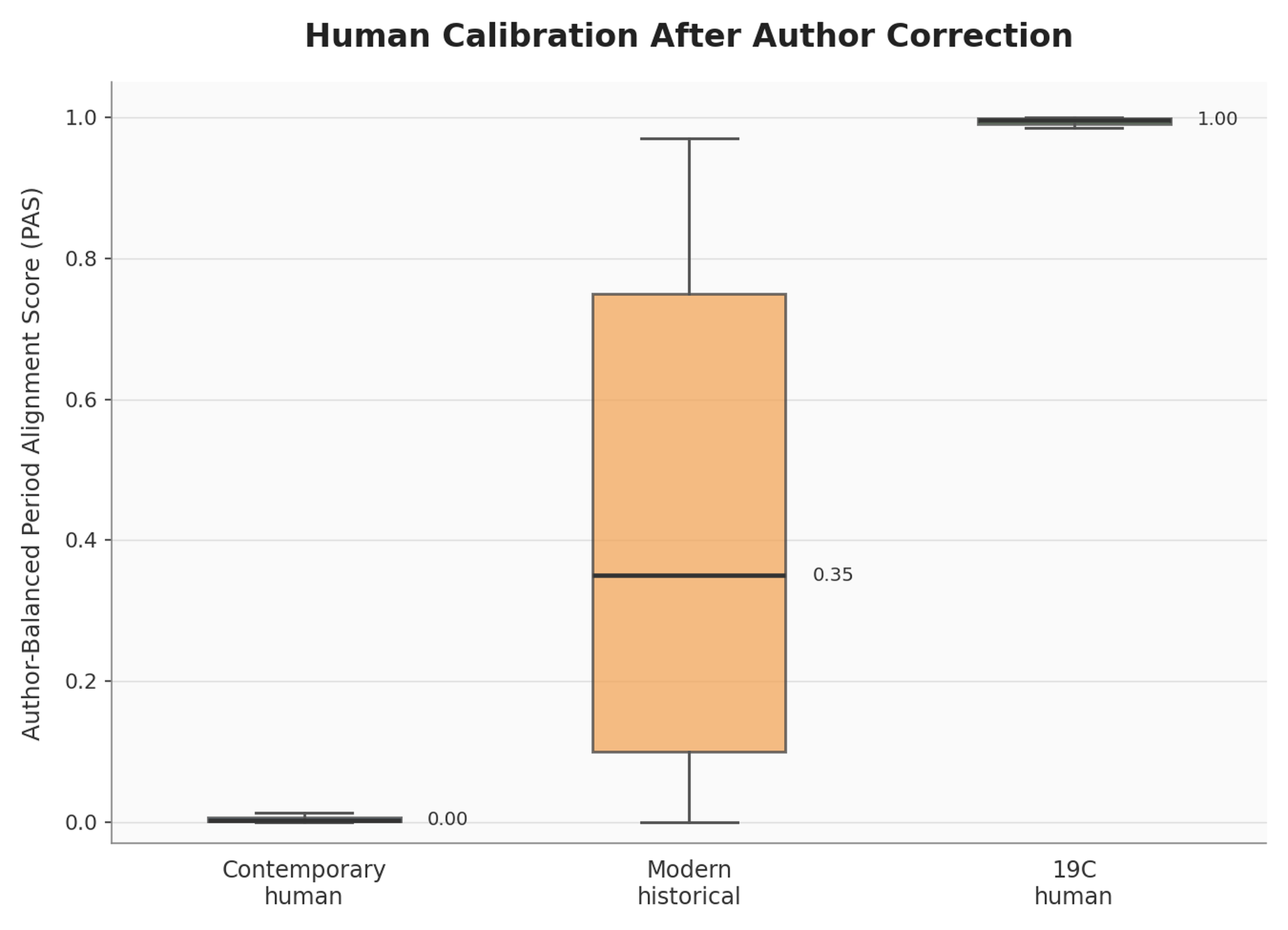}
\caption{Human calibration after author correction. Period Alignment Scores are summarized at the author level for the Zero-Style human comparator, modern historical fiction, and nineteenth-century fiction. Figure by the author, CC BY 4.0.}
\label{fig:1}
\end{figure*}

\subsection{Victorian Conditioning Produces Large Historical-Direction Shifts}

Across all three workflows, the novels generated with the Victorian instruction have much higher PAS values than the Zero-Style novels (Table~\ref{tab:3}). In GPT, the PAS value increases from .000032 to .02152 (Hedges's g = 1.215). In Qwen, this value goes from .000013 to .00599 (g = 1.101), and in Llama it increases from .01404 to .09270 (g = 1.233). Permutation p values are .0004 for GPT and Qwen and .0064 for Llama. Cliff's delta is 1.00 for GPT, .995 for Qwen, and .60 for Llama. The more difficult negative-group tests show that the GPT and Qwen shifts are not limited to the initial Zero-Style comparison. This robustness does not extend to the smaller Llama sample.

Under the primary calibration, the direction of movement is consistent across workflows and the standardized effects are large, yet the AI-generated texts remain near the modern end of this spectrum. Even Llama, with the highest average score in the Victorian condition among all the tested models, scores only .093, compared with .416 for modern historical human novels and .998 for nineteenth-century human works. These values are coordinates that depend on the comparison, not an absolute measure of historicity. Consequently, the most reasonable interpretation is not that AI-generated Victorian novels acquire historicity equal to that of nineteenth-century prose, but that the Victorian condition pushes the studied models in the historical direction under the primary calibration.

The HC3 regression reaches the same general conclusion. In the GPT workflow, treated as the reference, Zero-Style has a large negative coefficient relative to Victorian ($\beta$ = -5.09 for logit PAS, p < .001). The model $\times$ condition interaction indicates that the size of this difference varies in Qwen and Llama, while the logarithm of total word count is not significant (p = .609). This reduces concern about a simple text-length effect, but it cannot rule out effects of overall generation length on stylistic stability because generation length was not experimentally manipulated within each workflow.

The distinction between ``direction of movement'' and ``final destination'' persists when the bounded PAS probability is replaced with the normalized raw decision margin. On this scale, human Zero-Style authors define point 0 and nineteenth-century human authors define point 1; the average of modern historical human texts is .411. GPT goes from -.232 in the Zero-Style condition to .195 in the Victorian condition, Qwen from -.300 to .083, and Llama from .157 to .280. The same ordering is also seen in the Victorian-only model; there, the average of modern historical human texts is .421, and all three workflows again move toward the historical point, but still remain much lower than it. Therefore, the main interpretation is not an artifact of logistic-probability saturation, although this margin is still only a rescaling of the same classifier and is not considered an independent representation.

\begin{fullwidthtable}[htbp]
\centering
\begin{tabular}{@{}p{0.30\linewidth}rp{0.28\linewidth}p{0.23\linewidth}@{}}
\toprule
Group & Mean PAS & Comparison/effect & 95\% CI / p \\
\midrule
Human nineteenth-century (author mean) & .9981 & -- & -- \\
Modern historical human (author mean) & .4161 & -- & -- \\
Human Zero-Style comparator (author mean) & .0020 & -- & -- \\
GPT Victorian & .02152 & g = 1.215 vs Zero-Style & perm. p = .0004 \\
GPT Zero-Style & .000032 & -- & -- \\
Qwen Victorian & .00599 & g = 1.101 vs Zero-Style & perm. p = .0004 \\
Qwen Zero-Style & .000013 & -- & -- \\
Llama Victorian & .09270 & g = 1.233 vs Zero-Style & perm. p = .0064 \\
Llama Zero-Style & .01404 & -- & -- \\
Qwen paired shift & +.00598 & dz = .794 & 95\% CI [.00296, .00934]; p = .0001 \\
\bottomrule
\end{tabular}
\caption{Primary Period Alignment Scores and within-workflow comparisons.}
\label{tab:3}
\end{fullwidthtable}

\begin{figure*}[htbp]
\centering
\includegraphics[width=0.95\linewidth]{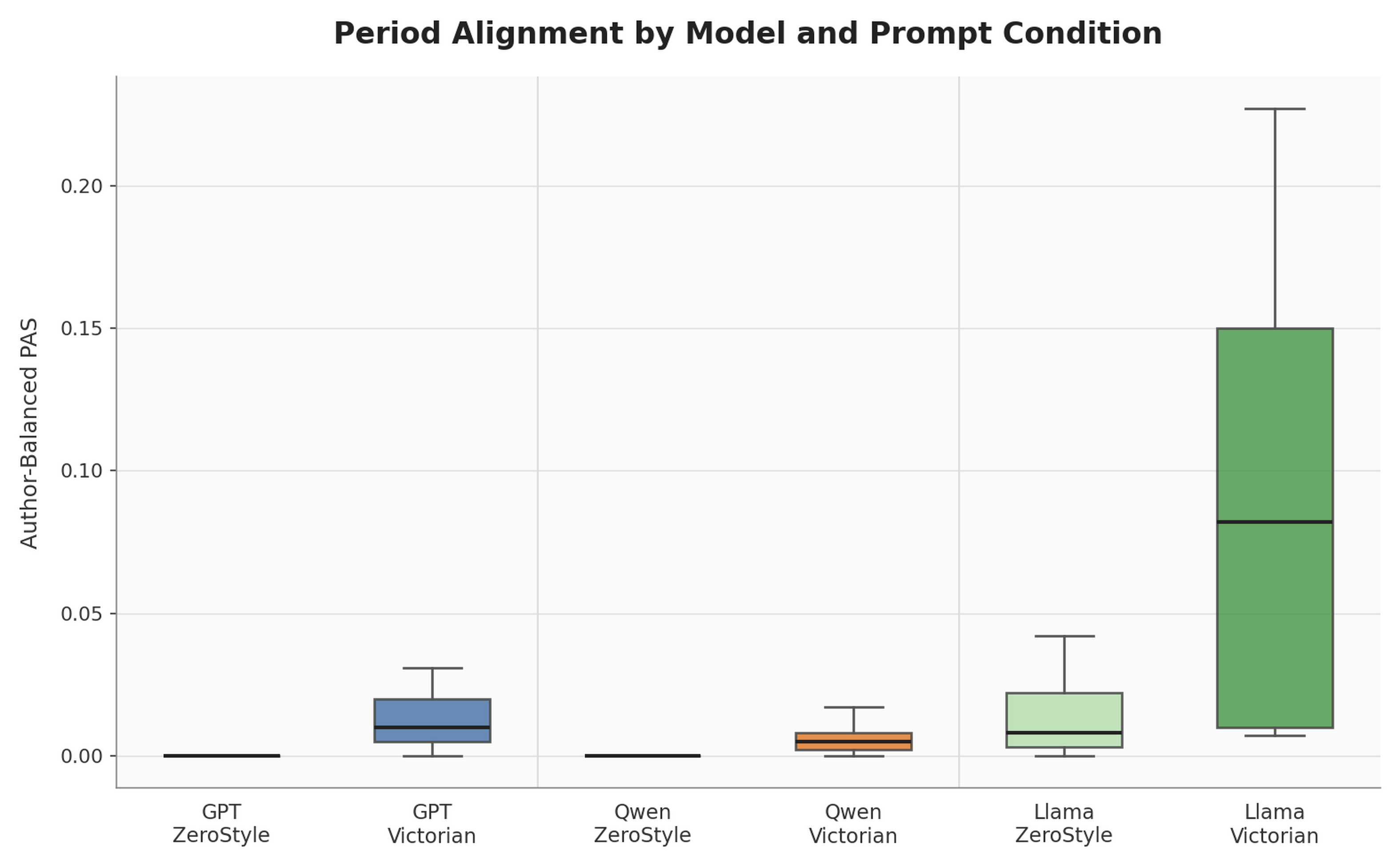}
\caption{Period Alignment Score by AI workflow and prompt condition after author-balanced human calibration. Figure by the author, CC BY 4.0.}
\label{fig:2}
\end{figure*}

\subsection{Paired Qwen Premises Isolate the Period Condition}

The paired Qwen design provides the strongest internal test. In twenty paired story premises, the mean PAS for the Victorian condition is .00598 higher than that for the Zero-Style condition; the 95 percent bootstrap confidence interval is [.00296, .00934]. The paired effect size is dz = .794. For the sign-flip permutation test, p = .0001, and for the Wilcoxon signed-rank test, p < .00001. With the wider negative groups and contemporary historical novels, the paired effect remains in the range dz = .65--.71, with sign-flip p = .0001.

\begin{figure*}[htbp]
\centering
\includegraphics[width=0.90\linewidth]{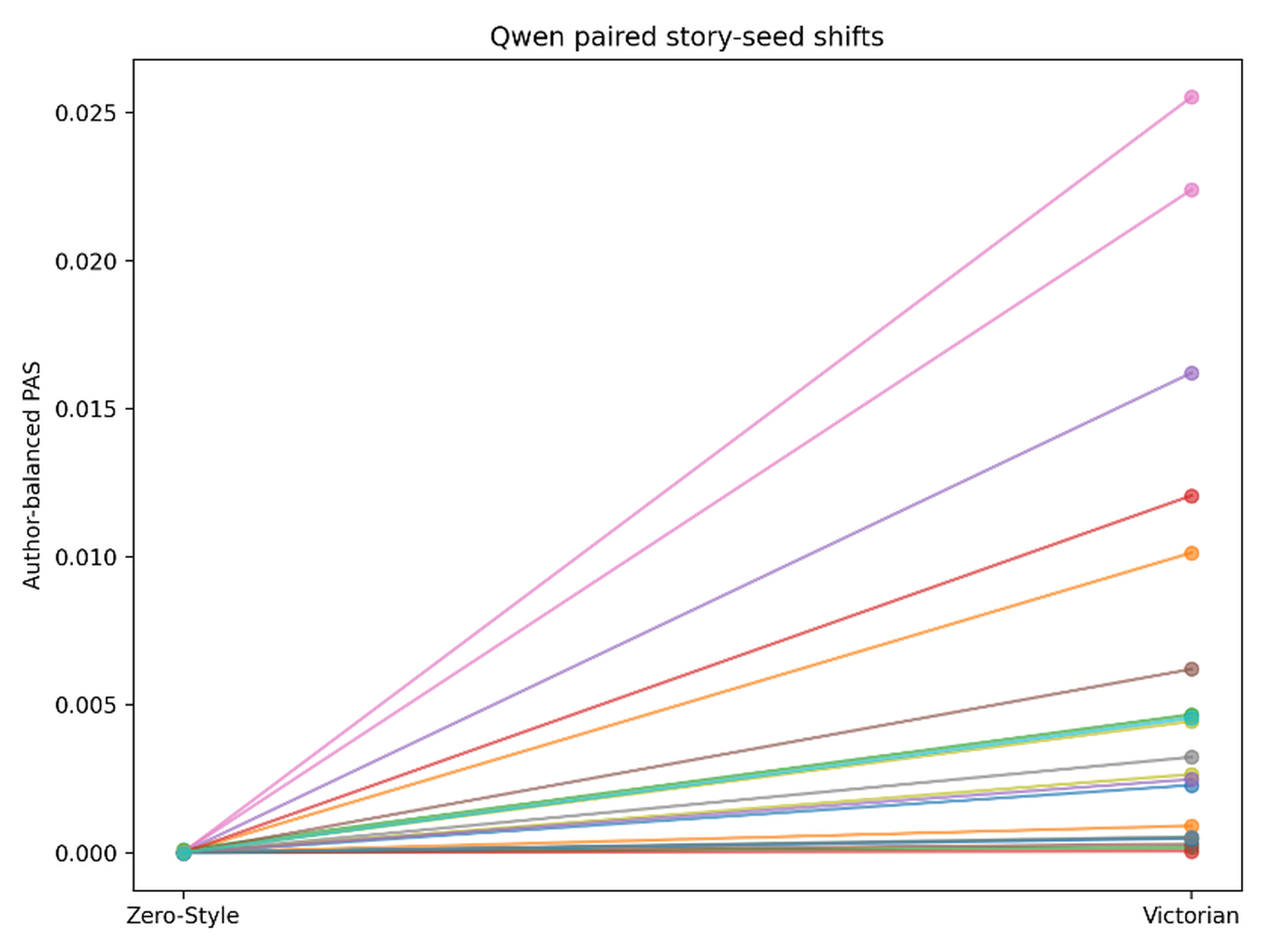}
\caption{Paired Qwen story-seed shifts. Each line connects the Zero-Style and Victorian realization of the same underlying narrative premise. Figure by the author, CC BY 4.0.}
\label{fig:3}
\end{figure*}

The plot of the individual pairs shows that this effect was not driven by only one or two extreme novels (Figure~\ref{fig:3}). Because each pair starts from the same story premise, this result limits many alternative explanations related to topic or plot. The Victorian instruction changes the formal realization of the same story idea. However, this design does not eliminate all differences between the prompts, because the Victorian and Zero-Style prompts necessarily specify different temporal and stylistic constraints. Therefore, the paired result should be considered as evidence for the effect of the ``period condition'' as a collection of instructions, not as the separate causal effect of only the word Victorian.

\subsection{The Historical Shift Is Layered Across Linguistic Levels}

Models constructed on the basis of different groups of features are hierarchically ordered (Table~\ref{tab:4}). The greatest and most stable effect of the condition can be observed in PAS when the model is trained using only function words. The effect of the difference between the Victorian and Zero-Style conditions is g = 5.02 for GPT, 1.76 for Qwen, and 1.81 for Llama. This is notable because function words are not obvious indicators of historical literature. They are simply frequent grammatical choices that emerge through tens of thousands of words. It would be misleading to interpret the larger GPT effect as evidence of superior model capability because the three workflows differ in prompting, generation guidance, decoding visibility, interface conditions, and generated text length. Equal 20,000-word samples standardize feature extraction, not the generation process itself.

POS organization follows the same pattern but to a lesser degree: g equals .65 for GPT, .53 for Qwen, and .70 for Llama. For syntax, the findings are less consistent. GPT demonstrates a substantial historical-direction effect (g = 1.18), Qwen a smaller one (g = .48), and Llama a small effect in the opposite direction (g = -.25).

What emerges from this is that the notion of ``Victorian style'' should not be treated as an absolute control variable. The prompt produces a set of effects whose transferability varies across the hierarchy of representational layers. Function-word patterns are highly responsive to prompting; grammatical-role distributions are mutable, but to a lesser degree; syntactic structure is workflow-dependent. Generative periodization is therefore layered within language.

\begin{table*}[htbp]
\centering
\begin{tabular}{@{}lrrr@{}}
\toprule
Feature family & GPT & Qwen & Llama \\
\midrule
Function words & 5.020 & 1.765 & 1.811 \\
POS + POS bigrams & .653 & .535 & .705 \\
Syntax/grammar & 1.183 & .483 & -.251 \\
All grammar features & 1.215 & 1.101 & 1.233 \\
\bottomrule
\end{tabular}
\caption{Victorian-versus-Zero-Style effect sizes by feature family.}
\label{tab:4}
\end{table*}

\subsection{Cross-Model Transfer Reveals a Shared Direction Beneath Different Baselines}

Cross-model transfer clarifies the distinction between ``stylistic position'' and ``stylistic direction.'' Decision threshold values may not carry over between models; for instance, the original balanced accuracy for GPT $\rightarrow$ Qwen transfer is .525, while for GPT $\rightarrow$ Llama it is .500. But the ranking order remains consistently reliable. Raw AUC is 1.00 in five directions and .99 in GPT $\rightarrow$ Llama (Table~\ref{tab:5}). Bootstrap uncertainty varies in some directions but increases once the relatively small Llama corpus is included in the comparison; for instance, the 95 percent bootstrap interval for Qwen $\rightarrow$ Llama stretches from .43 to 1.00. Training-label permutation tests yield p values under .05 in five of the six directions, while for Llama $\rightarrow$ Qwen the p value is only slightly above the cutoff (.057). Thus, the point estimates provide convincing evidence of a strong ranking direction shared by the models, although the resampling results show that the near-perfect AUCs are not equally precise across transfers.

With model-centered standardization, transfer accuracy becomes higher, reaching .95 to 1.00. This standardization uses the unlabeled distribution of the workflow in question. Hence, these figures cannot be said to reflect ordinary out-of-domain predictions. However, they help show that, once general baseline differences are stripped away, the Victorian-to-Zero-Style transfer direction becomes very similar across workflows. Nevertheless, for statistical analysis, the primary focus remains on AUC.

This difference helps explain why analyses that pool all AI workflows can be misleading: between-workflow baseline differences can exceed within-workflow condition differences. Nonetheless, cross-model ranking shows that a consistent period-condition direction can coexist with substantial baseline differences among workflows. The bootstrap results also show why this directional claim should be stated with appropriate uncertainty, especially for the smaller Llama sample.

\begin{fullwidthtable}[htbp]
\centering
\begin{tabular}{@{}p{0.25\linewidth}rp{0.27\linewidth}r@{}}
\toprule
Train $\rightarrow$ test & Observed raw AUC & Bootstrap 95\% CI & Permutation p \\
\midrule
GPT $\rightarrow$ Qwen & 1.000 & [1.000, 1.000] & .033 \\
GPT $\rightarrow$ Llama & .990 & [.860, 1.000] & .003 \\
Qwen $\rightarrow$ GPT & 1.000 & [1.000, 1.000] & .024 \\
Qwen $\rightarrow$ Llama & 1.000 & [.430, 1.000] & .002 \\
Llama $\rightarrow$ GPT & 1.000 & [.857, 1.000] & .003 \\
Llama $\rightarrow$ Qwen & 1.000 & [1.000, 1.000] & .057 \\
\bottomrule
\end{tabular}
\caption{Off-diagonal cross-model transfer with resampling uncertainty.}
\label{tab:5}
\end{fullwidthtable}

\subsection{Human Period Markers: Shared but Incomplete Alignment}

The author-balanced human comparison identifies 174 stable period markers. Of these, 78 (44.8\%) move in the historically expected direction in all three AI workflows, and 136 (78.2\%) move in that direction in at least two workflows. These numbers are considerably greater than the expected value in the permutation test with correlated features: for alignment in all three workflows, the null mean is 21.6 markers (95\% random interval: 6 to 53; p < .0001), and for alignment in at least two workflows, the null mean is 86.5 (95\% interval: 46 to 127; p = .0026). Alignment in each workflow is also greater than the expected value in permutation: GPT is aligned with 139 markers (p = .015), Qwen with 130 markers (p = .022), and Llama with 110 markers (p = .031). Shared shifts include features related to shall, may, must, which, nor, of, and several part-of-speech patterns and grammatical structures.

\section{Discussion}

\subsection{Historical Movement Is Not Historical Equivalence}

The central interpretive result is the distinction between ``direction of movement'' and ``final destination.'' The Victorian condition, within each generation workflow, produces large standardized shifts. However, once the same outputs are placed on the primary PAS calibration, AI Victorian means remain between .006 and .093, compared with .416 for modern historical human texts and .998 for nineteenth-century human texts. These figures do not constitute a scale of historicity; they show the position of the texts on the comparison scale on which the model was trained. Moreover, the decision margin-based sensitivity analysis again demonstrates the same ordering of these values on an unbounded version of the same scale. Here, modern human authors define 0 and nineteenth-century human authors define 1. The mean value of modern historical human texts is .411, while GPT, Qwen, and Llama have .195, .083, and .280, respectively.

This asymmetry helps reconcile apparently conflicting claims about LLM stylistic imitation. Models can respond strongly to stylistic prompts while remaining far from human exemplars. \textcite{mikros2025}, \textcite{bhandarkar2024}, and \textcite{wang2025} all report forms of partial rather than complete author-style reproduction. At the period level, the same pattern is complicated by the internal heterogeneity of the target category.

The modern historical corpus makes this clear. Modern historical novelists are not Victorian novelists, yet their grammatical patterns approach the historical pole of the primary PAS contrast considerably more than the AI-generated Victorians do. This accords with an important emphasis in neo-Victorian studies: modern historical fiction constructs ``the Victorians'' retrospectively rather than simply recapturing the past \parencite{heilmannllewellyn2010,hadley2010}. The comparator is not presented as a neo-Victorian genre sample, yet retrospective historical writing can reach higher levels of grammatical alignment in this calibration.

\subsection{Generative Periodization Is Layered}

The feature-family analysis reveals what aggregate PAS alone would hide: historical movement is unevenly distributed across linguistic levels. Function words are more responsive than POS groupings, while syntactic effects are more workflow-specific. This hierarchy matters because it shows that ``Victorian style'' is not a single uniformly controllable variable.

Function-word effects should not be considered superficial. In stylometry, function words operate below overt topic and register frequent patterns of grammatical choice \parencite{burrows2002,stamatatos2009}. The GPT function-word effect is large (g = 5.02), but its larger size relative to Qwen and Llama should not be interpreted as evidence of superior model capability because model identity is confounded with workflow differences. The 20,000-word sampling procedure equalizes the amount of text used for feature extraction, but it does not eliminate differences in total generation length, orchestration, or the opaque ChatGPT interface. The full-text word-count variable is not significant in the HC3 sensitivity analysis, but this observational check cannot rule out workflow effects associated with the generation of longer GPT novels. The function-word result is therefore not a direct demonstration of an abstract ``period grammar'': memorization, broader distributional generalization, or some combination of the two could contribute.

This selective productivity is captured by the concept of generative periodization, which does not specify the learning process involved. The period label is generative because it can produce uneven shifts across representational levels: some transfer across workflows, while others depend on the generation context.

\subsection{Shared Direction, Model-Mediated Realization}

In light of the cross-model findings, any assertion that each model has a unique manifestation of Victorian style becomes problematic. AUC values of .99--1.00 across the six transfer directions imply that the discrimination of the condition learned in one workflow well-orders the documents of another. However, the bootstrap confidence intervals are not identical in their precision, especially for the Llama 10-per-condition corpus, and one of the training-label permutations falls just above the conventional significance threshold at p = .057. The shared-direction claim should therefore be understood as strong descriptive and mostly inferential evidence, not as proof of universal transfer.

The strongest interpretation is therefore neither universality nor arbitrariness. Across the three workflows, the results indicate a shared directional signal in feature space associated with Victorian conditioning, but its realization remains model-mediated. This formulation avoids attributing human-like understanding to the models.

For computational literary studies, these findings suggest a more general method for examining categories used in prompts. If a cultural label produces recognizable differences within one model but those differences do not transfer to others, the label may be implemented through workflow-specific cues. If the differences persist across models after baseline normalization, the category may have a more stable and transferable operational signature.

The same logic can be applied to categories such as modernist, Gothic, realist, postcolonial, or noir. Each, however, should be measured and calibrated according to its own historical context; one fixed feature set should not be assumed to operate identically across literary traditions and periods.

More broadly, generative periodization provides a repeatable framework for examining how cultural and literary categories are transformed into operational patterns and constraints during text generation. From this perspective, Victorian is a case study rather than the endpoint of the method.

\subsection{Limitations}

Several limitations constrain the claims. First, the three AI corpora were generated under different technical conditions, so differences between models cannot be attributed solely to architecture or capability. Second, the historical calibration captures a broader nineteenth-century grammatical direction rather than a uniquely Victorian one; nevertheless, recalibration on 1837--1900 fiction preserves the main effects and remains highly accurate (author-level AUC = .9992). Third, each model-condition cell contains only one generation per novel design, limiting our ability to separate premise effects from stochastic generation variation; however, the very large GPT and Qwen effects suggest that the findings are not driven by outliers.

PAS is also contrast-dependent and should be interpreted as a relative coordinate rather than an absolute measure of historicity; alternative comparison corpora preserve the GPT and Qwen effects but weaken Llama. Finally, this study measures grammatical alignment rather than overall historical authenticity. Moreover, the transfer test is not directly deployable to unseen models, and unknown training-data provenance prevents us from distinguishing memorized historical patterns from broader learned representations. The claim is therefore behavioral: Victorian prompting reliably shifts historically calibrated grammatical patterns, especially for GPT and Qwen, without establishing how those patterns were learned.

\section{Conclusion}

Generative AI changes the usual relationship between text and historical period. In the past, labels such as ``Victorian'' were used after works were written to categorize them, but now these same labels can enter the prompt before text production and affect how it is formed.

The analysis of 100 novels shows that the Victorian prompt in GPT and Qwen clearly moves the texts toward nineteenth-century linguistic patterns. This result also persists across several robustness tests, although the findings show that this direction is broadly nineteenth-century rather than uniquely Victorian. GPT and Qwen also preserve their strong effects in harder tests, but the Llama result is weaker and more uncertain. The transfer of this pattern between models is also very strong, although uncertainty is greater in tests involving Llama.

These findings indicate that ``generative periodization'' can be a useful concept for computational literary studies. A historical label such as ``Victorian'' can affect text production; yet the label itself, the historical criteria of the study, and the features of the resulting text are not fully identical. The models do not fully reconstruct Victorian literature; instead, they generate reproducible changes whose meaning depends on the prompt and the comparison corpus.

The results offer a new perspective on literary history. Instead of asking only which features make existing texts Victorian, one may ask which features emerge when artificial intelligence is asked to produce ``Victorian.'' Thus, literary periods serve not only as classification schemes for the past but also as instructions for the future.

\section{Data and Code Availability}

Replication materials for this study are available through the OSF view-only link \url{https://osf.io/2wdtv/overview?view_only=3818049e4b634e9abaf0a0ec20b9be43}. The package includes the 410-text corpus list, bibliographic metadata, author-cleaning audit, 286-feature derived tables, pair IDs, robustness and sensitivity analyses, and the analysis notebooks used for feature extraction, author-balanced PAS estimation, paired and cross-model analyses, Victorian-boundary tests, and negative-class robustness analyses. Copyrighted human source texts are not redistributed; the package provides title-level metadata and derived features, while the historical corpus is publicly available from its source project.

\section*{Acknowledgements}

Generative AI tools were used to generate all analysis code. The author is responsible for the study design, validation of the analyses, interpretation of the results, and the claims reported in the article.

\printbibliography

\end{document}